\pdfoutput=1
\documentclass[11pt]{article}
\usepackage{acl}
\usepackage{times}
\usepackage{latexsym}
\usepackage[T1]{fontenc}
\usepackage[utf8]{inputenc}
\usepackage{microtype}
\usepackage{inconsolata}
\usepackage{graphicx}
\usepackage{stfloats}
\usepackage{amsmath}
\usepackage{amssymb}
\usepackage{amsfonts}
\usepackage{multirow}
\usepackage[normalem]{ulem}
\useunder{\uline}{\ul}{}

\title{SEAP: Training-free Sparse Expert Activation Pruning Unlock the Brainpower of Large Language Models}

\author{\textbf{Xun Liang}$^{1,}$\thanks{Equal contribution}~~~~\textbf{Hanyu Wang}$^{1,*}$~~~~\textbf{Huayi Lai}$^{1}$~~~~\textbf{Simin Niu}$^{1}$~~~~ 
\textbf{Shichao Song}$^{1}$~~~~\\\textbf{Jiawei Yang}$^{1}$~~~~\textbf{Jihao Zhao}$^{1}$~~~~\textbf{Feiyu Xiong}$^{2}$~~~~\textbf{Bo Tang}$^{2}$~~~~\textbf{Zhiyu Li}$^{2,}$\thanks{Corresponding author: lizy@iaar.ac.cn} \\
    $^1$School of Information, Renmin University of China, Beijing, China \quad \\
    $^2$Institute for Advanced Algorithms Research, Shanghai, China
}

\begin{document}
\maketitle
\begin{abstract}
Large Language Models have achieved remarkable success across various natural language processing tasks, yet their high computational cost during inference remains a major bottleneck. This paper introduces Sparse Expert Activation Pruning (SEAP)\footnote{Our code is available at \url{https://github.com/IAAR-Shanghai/SEAP}}, a training-free pruning method that selectively retains task-relevant parameters to reduce inference overhead. Inspired by the clustering patterns of hidden states and activations in LLMs, SEAP identifies task-specific expert activation patterns and prunes the model while preserving task performance and enhancing computational efficiency. Experimental results demonstrate that SEAP significantly reduces computational overhead while maintaining competitive accuracy. Notably, at 50\% pruning, SEAP surpasses both WandA and FLAP by over 20\%, and at 20\% pruning, it incurs only a 2.2\% performance drop compared to the dense model. These findings highlight SEAP’s scalability and effectiveness, making it a promising approach for optimizing large-scale LLMs.
\end{abstract}

\section{Introduction}
\label{sec:intro}

Large Language Models (LLMs) have achieved remarkable success across a wide spectrum of natural language processing (NLP) tasks \cite{LLMSurvey_Zhao_arxiv23,zheng_attention_2025}, demonstrating their versatility and adaptability in diverse applications. However, their deployment in real-world scenarios remains a significant challenge due to the substantial computational demands during inference. The inference process of LLMs is constrained by memory bandwidth and hardware limitations \cite{FLLLM_Chavan_IJCAI24}, making efficient deployment particularly difficult, especially in resource-constrained environments such as real-time systems and edge computing. As LLMs continue to scale, these challenges become even more pronounced, necessitating novel approaches to optimize computational efficiency while preserving model performance.

To mitigate the computational overhead of LLMs, several techniques have been explored. Quantization methods \cite{BinaryBERT_Bai_ACL21,OPTQ_Frantar_ICLR23} reduce weight precision, while Mixture of Experts (MoE) architectures \cite{MoE_Shazeer_ICLR17,Gshard_Lepikhin_arxiv20,SwitchTransformer_Fedus_JMLR22} dynamically activate only subsets of the network to improve efficiency. Another widely adopted approach is pruning \cite{SparseGPT_Frantar_arxiv23,LlmPruner_Ma_NeurIPS23,TEAL_liu_arxiv24}, which removes redundant parameters, neurons, or connections to reduce inference costs and storage requirements. Despite the effectiveness of pruning in reducing model complexity, most existing methods are static, relying on activation distributions collected from general datasets such as WikiText-2 \cite{wikitext2_Merity_arxiv16} and C4 \cite{c4_Raffel_JMLR20}. These methods apply a uniform pruning strategy across all tasks, which may lead to suboptimal efficiency and fail to fully leverage task-specific knowledge requirements.

Inspired by cognitive neuroscience, where different brain regions are selectively activated based on task demands, we hypothesize that a similar mechanism exists in LLMs—where different tasks rely on distinct sets of neurons working collaboratively. This suggests that pruning strategies should be adaptive rather than static, dynamically selecting the most relevant parameters for each task. By leveraging task-specific activation patterns, we can develop a more effective sparsification technique that maintains task performance while significantly enhancing computational efficiency.

\begin{figure*}[!ht]
    \centering
    \setlength{\belowcaptionskip}{0pt}
    \includegraphics[width=1\linewidth]{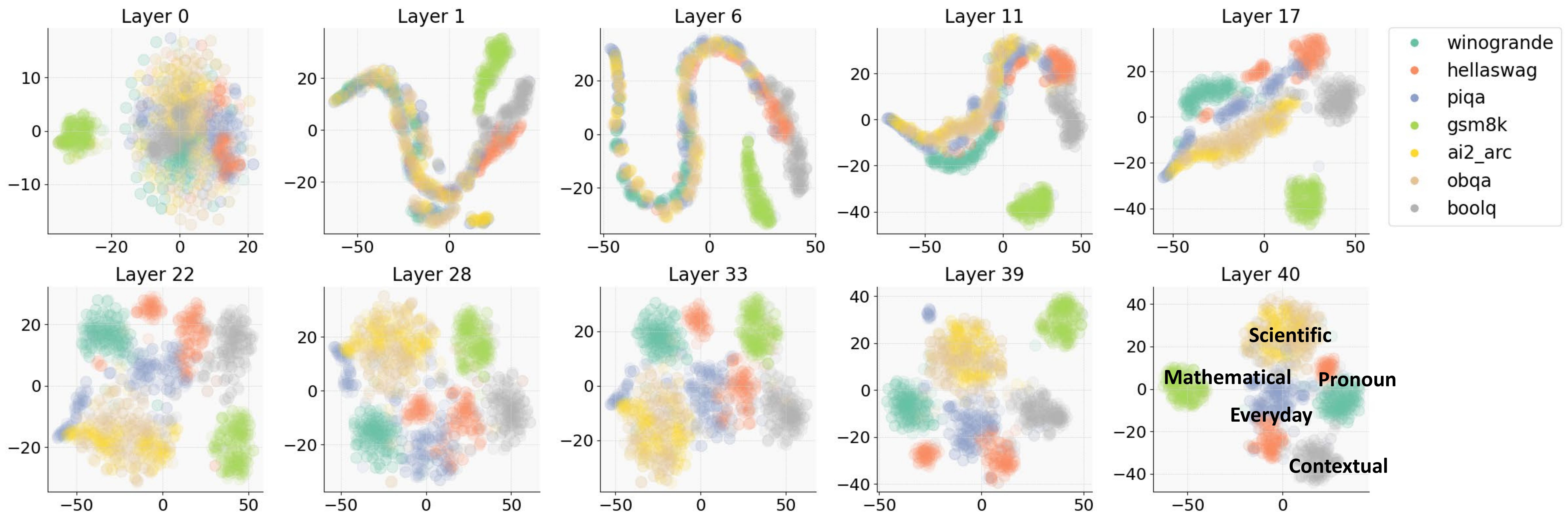}
    \caption{Visualization of hidden states $h(P)$ from different tasks. Each point represents the activation of a hidden state in the model for a specific task. The clustering patterns illustrate how tasks with similar requirements tend to activate similar regions in the model.}
    \label{fig:hidden_status}
\end{figure*}

\paragraph{Motivation Discovery}
\label{subsec: motivation}

In cognitive neuroscience, the brain parcellation theory posits that different regions of the brain are selectively activated based on specific task demands, thereby optimizing cognitive efficiency \cite{parcellation1_Mesulam_2000,parcellation2_Li_2022}. Inspired by this principle, we investigate whether a similar mechanism exists in LLMs — where distinct tasks may activate different sets of neurons, forming task-specific computational pathways. This perspective challenges conventional pruning approaches, which typically apply a uniform sparsity pattern across all tasks, potentially overlooking task-dependent knowledge representations.

We hypothesize that the knowledge requirements and activation patterns of different tasks are closely linked. If so, pruning should not be a one-size-fits-all process but rather a dynamic, task-aware strategy. By leveraging this relationship, pruning can adaptively retain the most relevant parameters based on each task’s specific characteristics, thereby enhancing computational efficiency while preserving task performance.

To validate this hypothesis, we design a multi-task experiment to analyze whether different tasks induce partitioned representations in LLM hidden states. We select seven task categories, covering a broad spectrum of linguistic and reasoning challenges, including common sense reasoning, mathematical problem-solving, and scientific question answering. We construct task-specific knowledge corpora consisting of question-answer pairs and feed them into an LLM to extract hidden states across multiple layers. The details of the corpus construction process are provided in Section~\ref{sec: appendix_exp}. To visualize the structure of these hidden states, we project their high-dimensional representations onto a two-dimensional plane. As shown in Figure~\ref{fig:hidden_status}, embeddings are initially intermingled. However, as forward propagation progresses, the model refines the semantic features of the input, leading to increasingly distinct task-specific clusters.

For instance, in the final layer, GSM8K\cite{gsm8k_Cobbe_arxiv21}, a challenging mathematical reasoning task, exhibits a clear separation from common-sense reasoning tasks. Similarly, OBQA\cite{obqa_Mihaylov_EMNLP18} and ARC\cite{arc_Peter_arxiv18}, both of which rely heavily on external scientific knowledge, form a closely related distribution. On the other hand, PIQA\cite{piqa_Bisk_AAAI20} and HellaSwag\cite{hellaswag_zellers_ACL19}, though both categorized as common-sense reasoning tasks, emphasize everyday knowledge, positioning them below OBQA and ARC in the visualization. Interestingly, Winogrande\cite{WinoGrande_Sakaguchi_arxiv19}, a pronoun resolution task, clusters with certain HellaSwag prompts, likely due to their shared reliance on pronoun-based reasoning. Lastly, BoolQ\cite{boolq_clark_NAACL19}, a contextual reasoning task, forms a distinct grouping, indicating its unique reliance on contextual comprehension.

\begin{figure*}[!ht]
    \centering
    \setlength{\belowcaptionskip}{0pt}
    \includegraphics[width=1.0\linewidth]{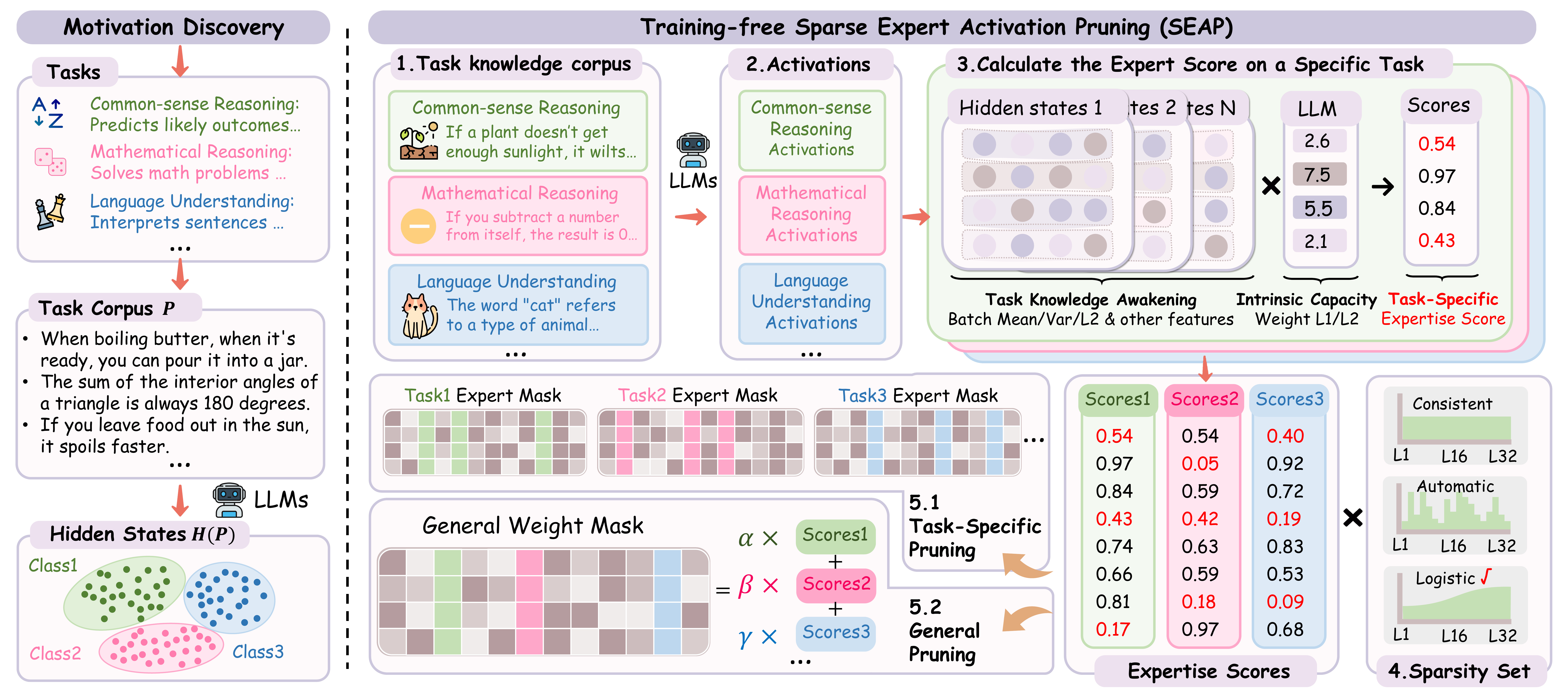}
    \caption{Framework of the SEAP approach. The \textbf{left} side shows the \textbf{Motivation Discovery} phase, where task-specific activation patterns are identified by analyzing hidden states and neuron activations extracted from the task corpus. The \textbf{right} side illustrates the \textbf{Training-free Sparse Expert Activation Pruning} process, consisting of five main steps described in Section \ref{subsec: SEAD}.}
    \label{fig:framework}
\end{figure*}
These findings suggest that each task occupies a distinct region within the hidden state space, with certain dimensions of activation corresponding to task-specific information. This observation draws an intriguing parallel to the functional specialization of the human brain, where different cognitive processes activate distinct neural circuits.

Building on this insight, we propose our central hypothesis: During inference, leveraging task-specific activation patterns and dynamically selecting the most relevant parameters can significantly reduce computational overhead while maintaining task performance. This task-adaptive pruning paradigm stands in contrast to traditional static pruning approaches, offering a promising path toward more efficient and specialized LLM deployment.

\paragraph{Contributions} Our key contributions are: 

\begin{itemize}
    \item We analyze task-specific activation patterns in LLMs, revealing their correlation with hidden state distributions and providing new insights for adaptive sparsification.
    \item We propose SEAP, a training-free, task-adaptive pruning method that dynamically adjusts sparsity based on task type, improving efficiency while preserving performance.
    \item We demonstrate that SEAP outperforms existing baselines in task accuracy, storage efficiency, and inference speed, confirming its effectiveness for efficient LLM deployment.
\end{itemize}

\section{Method}
\label{sec:method}

\subsection{Overview of SEAP}
\label{subsec: SEAD}

Building on the insights from Section~\ref{subsec: motivation}, we propose Sparse Expert Activation Pruning (SEAP), a training-free, task-adaptive pruning method that selectively activates task-relevant parameters during inference. By dynamically pruning based on task-specific activation patterns, SEAP reduces computational overhead while maintaining model performance. The SEAP Workflow (shown in Figure \ref{fig:framework}) is as follows,

\begin{enumerate}
    \item \textbf{Task-Specific Knowledge Corpus Construction}: We compile datasets from various tasks, such as reasoning, mathematical problem-solving, and scientific question answering, to form task-specific knowledge corpora (details in Section~\ref{sec: appendix_exp}).
    \item \textbf{Activation Patterns Modeling}: We feed the constructed corpora into an LLM and extract hidden state activations from multiple layers to analyze task-specific neural activity. This step lays the foundation for understanding how different tasks engage distinct parameter subsets.
    \item \textbf{Compute Neuron Importance Scores}: We perform task knowledge awakening by computing features such as mean, variance, and \(\ell\)2 norm from the collected activations, which are used to derive the task-specific expertise scores. These scores quantify the relevance of each neuron to the task and serve as the foundation for pruning decisions.
    \item \textbf{Distribute Sparsity Dynamically}: We introduce a logistic-based sparsity function that dynamically adjusts pruning ratios across layers, retaining critical neurons while maximizing efficiency. This enables structured sparsification tailored to task complexity.
    \item \textbf{Apply Task-Specific Pruning Strategies}: (5.1) Expert-Based Pruning: Task-specific expert scores are used to generate pruning masks, allowing the model to dynamically select the most relevant parameters during inference. (5.2) General Pruning: A unified pruning mask is created by aggregating scores across multiple tasks, ensuring broad applicability.
\end{enumerate}

\begin{figure*}[!hb]
    \centering
    \setlength{\belowcaptionskip}{0pt}
    \includegraphics[width=1.0\linewidth]{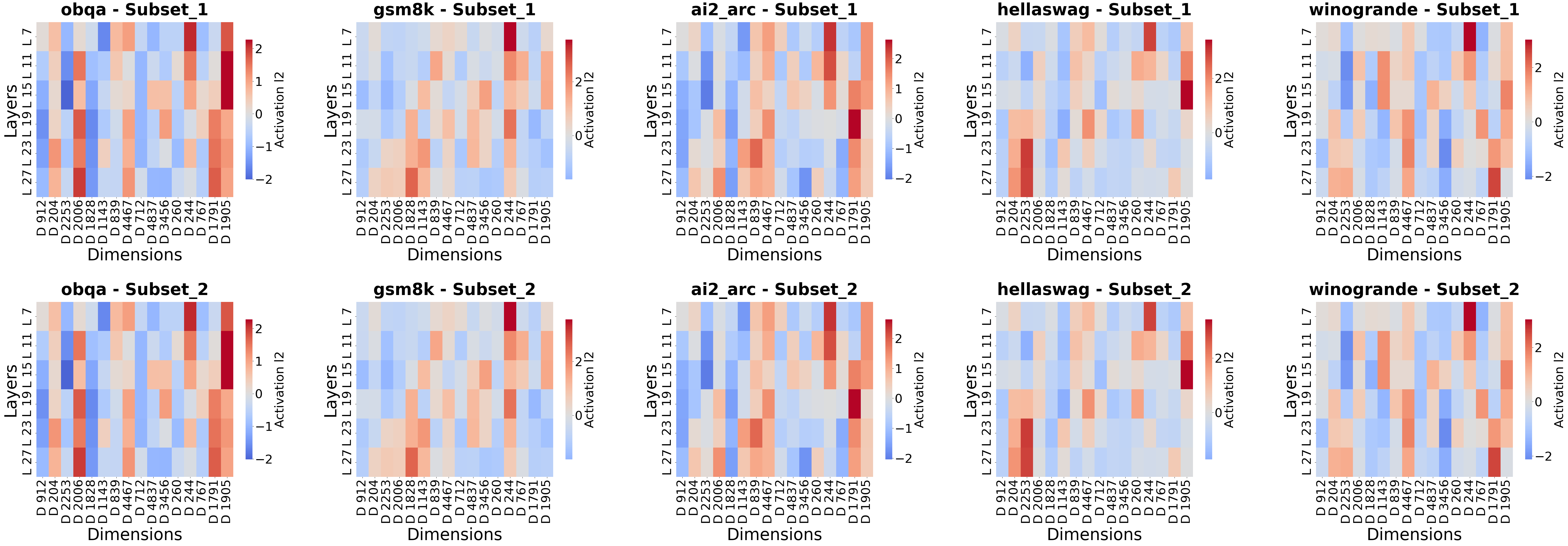}
    \caption{Heatmaps of dimension-wise average normalized \(\ell\)2 norms for different tasks. Each row corresponds to a layer or module, and each column represents a dimension in the hidden state space. The top and bottom parts of the figure show activation patterns from two randomly selected subsets of the same task. Consistent color patterns appear within tasks of the same type, while distinctly different tasks exhibit unique activation signatures, supporting our hypothesis that tasks selectively activate specific dimensions.}
    \label{fig:l2norms_heatmaps}
\end{figure*}

\subsection{Activation Patterns Modeling}
\label{subsec: activation_patterns}
To validate the feasibility of task-specific expert pruning, we analyze the consistency within task categories and the distinguishability between different task categories in the activations of LLMs. Only when both of these properties are present can the task-specific expert pruning method be effectively applied.
To formalize our analysis, let \( \tau \) denote a task type, and let \( p_i \) be a specific prompt within task \(\tau\).
We denote by
\begin{equation}
\label{eq:hidstates}
\mathbf{h}(p_i)^\tau 
= \bigl[h_1(p_i)^\tau,\,h_2(p_i)^\tau,\,\dots,\,h_C(p_i)^\tau \bigr]
\end{equation}
the hidden state vector of dimension \(C\) extracted from a particular layer of the model.
We further define 
\begin{equation}
\label{eq:hset}
H^\tau = \bigl\{
\mathbf{h}(p_1)^\tau,\,
\mathbf{h}(p_2)^\tau,\,
\dots,\,
\mathbf{h}(p_{n_\tau})^\tau
\bigr\},
\end{equation}
where \(n_\tau\) is the number of prompts for task \(\tau\).

To quantify how each dimension (often viewed as a “neuron channel”) responds under different tasks, we define the following statistical measures that capture the mean activation level, variance, and \(\ell\)2 norm of each dimension:
\begin{align}
\mu_j^\tau 
&= \frac{1}{n_\tau} \sum_{i=1}^{n_\tau} h_j\bigl(p_i\bigr)^\tau,
\label{eq:mean_activation}
\\[6pt]
(\sigma_j^\tau)^2 
&= \frac{1}{n_\tau}\sum_{i=1}^{n_\tau}\Bigl(h_j(p_i)^\tau - \mu_j^\tau\Bigr)^2,
\label{eq:var_activation}
\\[6pt]
\overline{\bigl\|h_j^\tau\bigr\|_2}
&= \frac{\bigl\|h_j^\tau\bigr\|_2}{n_\tau} 
= \frac{1}{n_\tau}\,\sqrt{\sum_{i=1}^{n_\tau}\Bigl(h_j(p_i)^\tau\Bigr)^2}.
\label{eq:l2norm_activation}
\end{align}
In these equations, \(\mu_j^\tau\) denotes the mean activation of dimension \(j\) for task \(\tau\), 
\((\sigma_j^\tau)^2\) represents its variance, and \(\bigl\|h_j^\tau\bigr\|_2\) is the total \(\ell\)2 norm of that dimension’s activations. 
The normalized measure \(\overline{\bigl\|h_j^\tau\bigr\|_2}\) divides the raw \(\ell\)2 norm by \(n_\tau\). 

Figure~\ref{fig:l2norms_heatmaps} visualizes \(\overline{\bigl\|h_j^\tau\bigr\|_2}\) for multiple tasks in a given layer or module. 
Columns represent different dimensions, while rows correspond to either distinct layers or modules. 
Bright regions indicate higher \(\ell\)2 norms, suggesting stronger activation in those dimensions.  
Within each task, two random subsets of prompts often reveal remarkably consistent high-activation dimensions, indicating internal stability. 
In contrast, tasks with very different objectives exhibit distinct “hot spots,” implying that they engage disjoint sets of dimensions. 
Hence, these dimension-wise patterns reinforce our claim: tasks of the same type focus on overlapping dimensions, whereas tasks from different domains activate largely separate regions of the hidden state.

Furthermore, consider the weight matrix \(W \in \mathbb{R}^{A \times C}\) that applies a linear transformation to the hidden state. 
We conceptualize \(W\) as consisting of \(C\) column “slices,” where each slice \(w_i \in \mathbb{R}^{A}\) corresponds to the \(i\)-th dimension in the input hidden state. 
If a particular dimension \(i\) is consistently unimportant for a given task \(\tau\), then its associated column \(w_i\) can be pruned—thereby reducing computation without sacrificing essential features.

\begin{figure}[!h]
    \centering
    \setlength{\belowcaptionskip}{0pt}
    \includegraphics[width=1.0\linewidth]{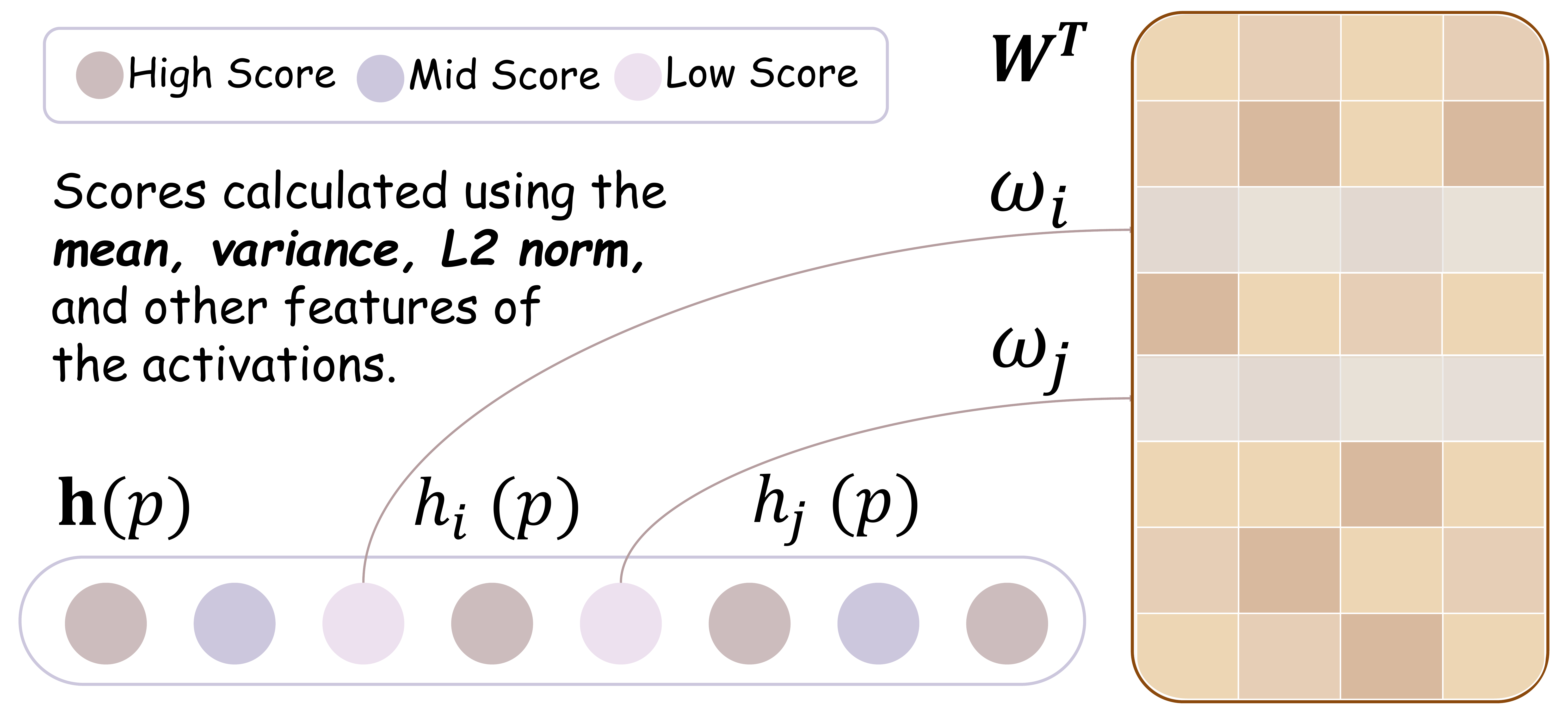}
    \caption{Illustration of how neurons are pruned based on importance scores.}
    \label{fig:calculate}
\end{figure}

Formally, for a prompt \(p_i\) in task \(\tau\), 
the output \(\mathbf{o} \in \mathbb{R}^A\) of the linear layer is
\begin{align}
\label{eq:linear_composition}
\mathbf{o}
&~=~W \,\mathbf{h}(p_i)^\tau \nonumber\\[6pt]
&~=~
\begin{bmatrix}
w_{1,1} & \cdots & w_{1,C} \\
\vdots  & \ddots & \vdots  \\
w_{A,1} & \cdots & w_{A,C}
\end{bmatrix}
\cdot
\begin{bmatrix}
h_1(p_i)^\tau\\
\vdots\\
h_C(p_i)^\tau
\end{bmatrix},
\end{align}
Here, \(w_{a,i}\) is the weight linking the \(i\)-th hidden dimension to the \(a\)-th output unit. 
If dimension \(i\) is deemed unimportant for task \(\tau\), we zero out the entire column 
\(\{w_{1,i}, w_{2,i}, \dots, w_{A,i}\}\) (see Figure~\ref{fig:calculate}), resulting in a form of structured sparsity that is hardware-friendly for inference acceleration.

In summary, by identifying and pruning inactive or low-importance dimensions on a per-task basis, we can achieve task-adaptive compression.

\subsection{Pruning Procedure}
\label{subsec: pruning_procedure}
As mentioned, the next step is to prune neurons based on their importance scores for each task. Specifically, the neuron importance \(s_i^{(\ell, \tau)}\) for each task is computed using a function \(f(\cdot)\), which combines the aforementioned statistical measures with the weight norm of the corresponding neuron in layer \(\ell\). Formally,
\begin{equation}
\label{eq:score_function}
s_i^{(\ell, \tau)} 
\;=\;
f\Bigl(
   \mu_i^{(\ell,\tau)},\;
   \bigl(\sigma_i^{(\ell,\tau)}\bigr)^2,\;
   \overline{\bigl\|h_i^{(\ell,\tau)}\bigr\|}_2,\;
   w_i^{(\ell)}
\Bigr),
\end{equation}

where \(\mu_i^{(\ell,\tau)}\) and \(\bigl(\sigma_i^{(\ell,\tau)}\bigr)^2\) denote the mean and variance of the activations for neuron \(i\) in layer \(\ell\) under task \(\tau\), \(\overline{\|h_i^{(\ell,\tau)}\|}_2\) represents the average \(\ell\)2 norm of its activations, and \(w_i^{(\ell)}\) is the corresponding weight in layer \(\ell\).

Once these importance values are obtained for all \(C\) neurons in the chosen module of layer \(\ell\), we collect \(\{|s^{(\ell, \tau)}_{1}|, \dots, |s^{(\ell, \tau)}_{C}|\}\) and sort them in ascending order:
\begin{equation}
\label{eq:sorted_scores}
|s^{(\ell, \tau)}|_{\mathrm{sorted}} 
\;=\; 
\mathrm{Sort}\!\Bigl(
   |s^{(\ell, \tau)}_{1}|,\,\dots,\,|s^{(\ell, \tau)}_{C}|
\Bigr).
\end{equation}

Given a desired sparsity ratio \(\rho \in [0,1]\), we identify the \(\rho\)-th quantile in \eqref{eq:sorted_scores}:
\begin{equation}
\label{eq:quantile_threshold}
\theta^{(\ell, \tau)} 
\;=\; 
|s^{(\ell, \tau)}|_{\mathrm{sorted}}\!\bigl(\lfloor \rho\,C \rfloor\bigr),
\end{equation}
where \(\lfloor \cdot \rfloor\) is the floor function.  
All neurons whose importance \(|s^{(\ell, \tau)}_{i}|\) is less than or equal to \(\theta^{(\ell, \tau)}\) are then pruned:
\begin{equation}
\label{eq:prune_rule}
w^{(\ell)}_{i}
\;=\;
\begin{cases}
0, & \text{if } |s^{(\ell, \tau)}_{i}|\leq \theta^{(\ell, \tau)} ,\\[6pt]
w^{(\ell)}_{i}, & \text{otherwise}.
\end{cases}
\end{equation}
Here, \(w^{(\ell)}_{i}\!=0\) effectively disables neuron \(i\).

\subsection{Scores Calculating Strategy}
\label{subsec: scores_strategy}
The expert activation pruning framework we propose is highly flexible, capable of accommodating various neuron importance metrics. The importance score \(s_i^{(\ell,\tau)}\) can be easily integrated with existing training-free methods, such as \textbf{FLAP}\cite{flap_An_AAAI24} and \textbf{WandA}\cite{wanda_Sun_ICLR24}, by simply adjusting their respective formulas.

In the framework above, each neuron \(i\) in layer \(\ell\) under task \(\tau\) is assigned an importance score \(s_i^{(\ell,\tau)}\) by combining its activation statistics and weight information. Specifically, we rely on the definitions of mean activation, variance, and total activation energy (i.e., squared \(\ell_2\)-norm) from equations 
\eqref{eq:mean_activation}, 
\eqref{eq:var_activation}, and 
\eqref{eq:l2norm_activation} in Section~\ref{subsec: activation_patterns}, respectively. 
Let \(w_i^{(\ell)} \in \mathbb{R}^{D_\ell}\) be the weight vector corresponding to neuron \(i\) in layer \(\ell\), with \(\|w_i^{(\ell)}\|_{2}\) and \(\|w_i^{(\ell)}\|_{1}\) denoting its \(\ell_2\)- and \(\ell_1\)-norms.

Based on these quantities, we introduce two specific scoring functions, \(s_F\) and \(s_W\). The first, \(s_F\), follows the scoring method used in FLAP by multiplying the neuron’s variance with the squared \(\ell_2\)-norm of its weights:
\begin{equation}
s_{F,i}^{(\ell,\tau)}
~=~
(\sigma_i^{(\ell,\tau)})^{2}
~\times~
\bigl\|w_i^{(\ell)}\bigr\|_{2}^{2},
\label{eq:score_sf}
\end{equation}
This gives higher importance to neurons whose activations vary significantly across prompts and whose weight magnitudes are relatively large.

The second, \(s_W\), follows the scoring method used in WandA, replacing the variance with the total activation energy (i.e., the squared \(\ell_2\)-norm of the neuron’s activations) and weighting it by \(\bigl\|w_i^{(\ell)}\bigr\|_{1}\):
\begin{equation}
s_{W,i}^{(\ell,\tau)}
~=~
\bigl\|h_i^{(\ell,\tau)}\bigr\|_{2}^{2}
~\times~
\bigl\|w_i^{(\ell)}\bigr\|_{1}.
\label{eq:score_sw}
\end{equation}

After computing \(s_F\) or \(s_W\) for all neurons, we apply the threshold-based pruning procedure described in Section~\ref{subsec: pruning_procedure} to remove those receiving lower scores. 

In the MLP layers, each neuron index \(i\) directly corresponds to a single channel,
so the scores \(s_F\) or \(s_W\) in \eqref{eq:score_sf}--\eqref{eq:score_sw} apply on a channel-by-channel basis.
By contrast, in the attention layers, we aggregate neuron scores at the head level:
suppose each attention head \(h\) spans a contiguous set of dimensions \(\mathcal{I}_h\), then its overall importance score can be taken as
\begin{equation}
s_{h}^{(\ell,\tau)} 
~=~ 
\sum_{i \,\in\, \mathcal{I}_h} s_{i}^{(\ell,\tau)}.
\end{equation}
A threshold is then applied to \(s_{h}^{(\ell,\tau)}\) to prune or retain the entire head. 

\subsection{Expert-Based vs General Pruning}
\label{subsec: pruning_strategies}
We propose two pruning strategies to adapt to task-specific scenarios and general scenarios.
The first strategy, \textbf{Expert-based Pruning}, uses task-specific importance scores to prune neurons or attention heads. The pruning process selects the importance score \( s_i^{(\ell)} \) for each neuron \( i \) in layer \( \ell \) based on the task type \(\tau_{\text{chosen}}\) as follows:
\begin{equation}
s_i^{(\ell)} = s_i^{(\ell,\tau_{\text{chosen}})},
\end{equation}
where \( \tau_{\text{chosen}} \) is the task selected for pruning, and \( s_i^{(\ell,\tau_{\text{chosen}})} \) is the corresponding importance score. During inference, different pruning masks can be flexibly applied based on the task type.

The second strategy, \textbf{General Pruning}, integrates importance scores across multiple tasks to identify neurons or attention heads that are less important across all tasks. This general pruning approach forms a unified model, ensuring that important components are retained across a broader range of tasks. The score is computed as a weighted average of the importance scores from each task:
\begin{equation}
s_i^{(\ell)} = \sum_{\tau} \alpha_{\tau} s_i^{(\ell,\tau)},
\end{equation}
where \( \alpha_{\tau} \) is the weight assigned to task \(\tau\), and the sum is taken across all tasks.

\subsection{Sparsity Setting}
\label{subsec: sparsity_setting}
\begin{figure}[ht]
    \centering
    \setlength{\belowcaptionskip}{0pt}
    \includegraphics[width=\linewidth]{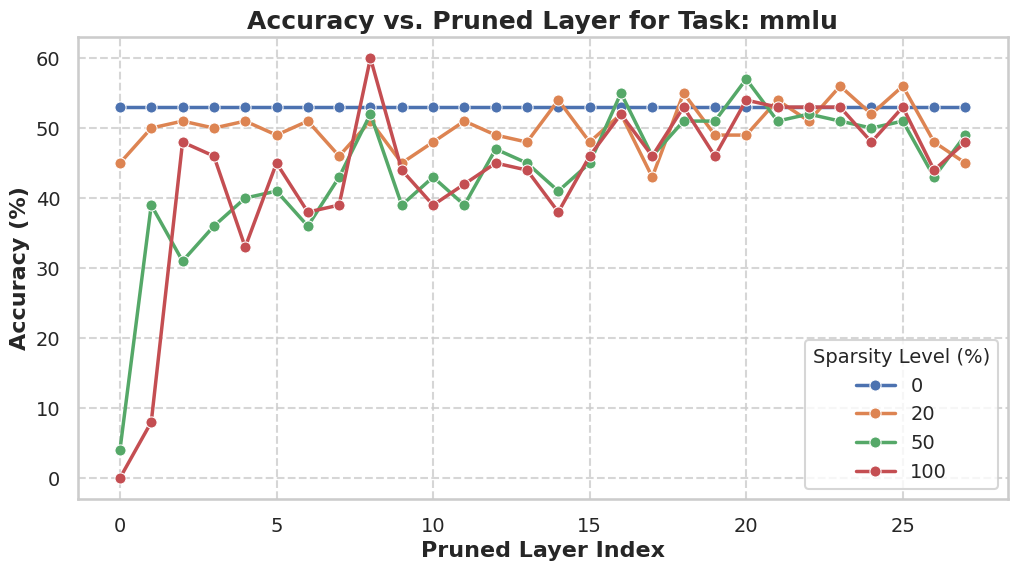}
    \caption{Impact of pruning on MMLU performance at different layers and sparsity levels. Early layers are more sensitive to pruning.}
    \label{fig:mmlu_rm_test}
\end{figure}

To determine appropriate sparsity levels for each layer in LLMs, we conduct a \emph{remove test} on two tasks: MMLU\cite{mmlu_Hendrycks_ICLR21} and PIQA. This test prunes neurons across layers at varying sparsity levels and mesures task performance. Figures~\ref{fig:mmlu_rm_test} and~\ref{fig:piqa_rm_test} show that early layers are more sensitive to pruning, while deeper layers tolerate higher sparsity with minimal performance loss, consistent with our observations in Section~\ref{subsec: motivation}.

Additionally, LLM-Pruner\cite{LlmPruner_Ma_NeurIPS23} and FLAP methods highlight that layers near the output are crucial for language modeling. Thus, we set the sparsity of the final \(n\) layers to zero and adjust the sparsity of other layers to maintain overall sparsity.
For sparsity setting, we employ a differentiable logistic function to ensure a smooth and continuous distribution of sparsity across layers. Each layer index \(\ell\) is mapped to the interval \([0, 1]\) using \(x_\ell = \frac{\ell - 1}{L - 1}\), where \(L\) is the total number of layers. The sparsity for layer \(\ell\) is defined as:
\begin{equation}
\label{eq:logistic_sparsity}
\rho_\ell 
~\;=\;
\rho\bigl(x_\ell\bigr)
~\;=\;
\Lambda\;\frac{1}{1 + \exp\bigl(-k(x_\ell - x_0)\bigr)},
\end{equation}
where \(k\) controls the steepness, \(x_0\) sets the inflection point, and \(\Lambda\) represents the maximum sparsity. This ensures lower sparsity in early layers and progressively higher sparsity in deeper layers.

To meet a global sparsity target \(G\), we adjust \(\Lambda\) so that the average sparsity satisfies:
\begin{equation}
\label{eq:avg_sparsity}
\frac{1}{L} \sum_{\ell=1}^{L}\rho_{\ell}
~=~ 
G.
\end{equation}
This is done via a numerical search for \(\Lambda\). In our experiments, we use \((x_0, k) = (0.3, 1)\).

\renewcommand{\arraystretch}{1.25}
\begin{table*}[ht]
\small 
\setlength{\belowcaptionskip}{0pt}
\begin{tabular}{clcccccccc}
\hline
\multirow{2}{*}{\begin{tabular}[c]{@{}c@{}}Pruning\\ Ratio\end{tabular}} & \multicolumn{1}{c}{\multirow{2}{*}{Method}} & \multicolumn{8}{c}{Llama-2-7B}                                                                                                        \\ \cline{3-10} 
                                                                         & \multicolumn{1}{c}{}                        & WinoGrande     & OBQA           & HellaSwag      & PIQA           & ARC-c          & ARC-e          & BoolQ          & Average        \\ \hline
0\%                                                                      & Dense                                       & 69.14          & 44.20          & 76.01          & 79.11          & 46.33          & 76.26          & 77.71          & 66.97          \\ \hline
\multirow{6}{*}{20\%}                                                    & WandA-sp (\(s_W\))                             & 62.67          & 40.80          & 71.56          & 76.28          & 42.41          & 71.93          & 61.01          & 60.95          \\
                                                                         & SEAP (\(s_W\))                                 & 66.77          & \textbf{43.00} & 72.53          & 77.80          & \textbf{45.48} & \textbf{75.42} & {\ul 71.77}    & 64.68          \\
                                                                         & SEAP-gen (\(s_W\))                             & {\ul 67.80}    & 41.00          & 73.77          & 77.58          & 44.71          & {\ul 74.33}    & 71.44          & 64.38          \\ \cline{2-10} 
                                                                         & FLAP (\(s_F\))                                 & 67.32          & 41.00          & 72.77          & 76.12          & 42.75          & 71.93          & 62.57          & 62.07          \\
                                                                         & SEAP (\(s_F\))                                 & \textbf{68.19} & {\ul 42.60}    & {\ul 74.07}    & {\ul 78.07}    & {\ul 45.39}    & \textbf{75.42} & \textbf{74.50} & \textbf{65.46} \\
                                                                         & SEAP-gen (\(s_F\))                             & 67.72          & 41.20          & \textbf{74.82} & \textbf{78.35} & {\ul 45.39}    & 74.12          & 71.68          & {\ul 64.75}    \\ \hline
\multirow{6}{*}{50\%}                                                    & WandA-sp (\(s_W\))                             & 52.72          & 35.20          & 41.11          & 64.36          & 30.97          & 52.78          & 39.45          & 45.23          \\
                                                                         & SEAP (\(s_W\))                                 & 56.12          & 37.20          & 58.07          & {\ul 73.83}    & \textbf{38.74} & \textbf{61.32} & \textbf{60.15} & {\ul 55.06}    \\
                                                                         & SEAP-gen (\(s_W\))                             & 54.70          & 38.20          & 56.97          & 71.76          & 35.24          & 57.15          & 57.25          & 53.04          \\ \cline{2-10} 
                                                                         & FLAP (\(s_F\))                                 & 56.04          & 34.40          & 48.62          & 63.00          & 32.17          & 51.18          & 42.32          & 46.82          \\
                                                                         & SEAP (\(s_F\))                                 & \textbf{60.14} & {\ul 38.80}    & \textbf{58.22} & \textbf{74.32} & {\ul 38.14}    & {\ul 60.56}    & {\ul 59.94}    & \textbf{55.73} \\
                                                                         & SEAP-gen (\(s_F\))                             & {\ul 59.91}    & \textbf{39.80} & {\ul 58.17}    & 73.39          & 37.97          & 55.72          & 57.98          & 54.71          \\ \hline
\end{tabular}
\caption{Task performance accuracy on Llama-2-7B under different pruning ratios. A higher ↑ score indicates better performance. The \textbf{bolded} entries represent the highest scoring methods, while the \underline{underlined} entries represent the second highest scoring methods.}
\label{tab:llama2_7b}
\end{table*}

\section{Experiment and Results Analysis}
\label{sec:experiment}
\subsection{Experimental Settings}
\label{subsec: experiment_settings}
\paragraph{LLMs and Tasks} 
We evaluate our method on the Llama2-7B and Llama2-13B models, assessing their performance across a range of downstream tasks. Zero-shot performance is evaluated on seven benchmarks—BoolQ, ARC Easy, ARC Challenge, HellaSwag, OBQA, PiQA, and Winogrande—using the EleutherAI LM Harness\cite{eval-harness}. In addition to accuracy, we also compare inference speed. More details can be found in Section~\ref{sec: appendix_exp}.

\paragraph{Baselines} 
We compare our method to the original (dense) models and two established training-free sparsification methods: WandA and FLAP. The key difference between these baselines is in the importance score calculation. For the expert-based and general models, we use the scoring methods \(s_W\) from WandA and \(s_F\) from FLAP, respectively. All methods, including baselines, adopt the logistic sparsity setting proposed in this paper, enabling consistent comparison of knowledge corpus expert activation differences. A detailed comparison of sparsity settings is provided in Section \ref{subsec: results}.

\subsection{Results and Analysis}
\label{subsec: results}
\paragraph{Zero-shot Tasks Performance}

We evaluate SEAP's zero-shot performance across multiple benchmarks, demonstrating its ability to reduce computational overhead while maintaining competitive accuracy. For Llama-2-7B (see Table~\ref{tab:llama2_7b}), at 20\% pruning, SEAP outperforms both WandA and FLAP with minimal performance loss, showing only a 2.2\% drop compared to the dense model, which is exceptional for structured pruning. At 50\% pruning, SEAP's advantage over FLAP and WandA increases, with the average score surpassing both baselines by over 20\%, indicating strong performance even at high sparsity levels.

Interestingly, the results do not always align with expectations for general versus expert models. In HellaSwag, the general model outperforms the expert model, likely due to richer knowledge corpora enhancing task-relevant activation distributions. A similar trend is observed in BoolQ with Llama-2-13B (see Table~\ref{tab:llama2_13b}), where higher sparsity leads to a noticeable performance drop, possibly due to the simpler True/False nature of the task, which lacks a sufficiently rich knowledge corpus for task-specific pruning to be fully effective.

Overall, our results confirm that task-specific pruning improves efficiency without compromising performance.

\paragraph{Inference Speed} 
Our pruning method completes pruning on Llama-2-7B in approximately 5–10 minutes on a single NVIDIA H800 80GB GPU. 
As shown in Table~\ref{tab:speed}, SEAP significantly improves inference speed compared to non-structured pruning methods like WandA. At 20\% pruning, SEAP is slightly slower than FLAP, but at 50\% pruning, SEAP maintains high speed with only a minimal difference compared to FLAP. 
These results demonstrate that SEAP reduces computational resources while maintaining high inference speed, making it suitable for real-world deployment across various hardware environments.

\renewcommand{\arraystretch}{1.25}
\begin{table}[h]
\setlength{\belowcaptionskip}{0pt}
\small
\begin{tabular}{llp{0.8cm}p{0.8cm}p{0.8cm}p{0.8cm}}
\hline
\multirow{2}{*}{Ratio} & \multirow{2}{*}{Method} & \multicolumn{2}{c}{Llama-2-7B}  & \multicolumn{2}{c}{Llama-2-13B} \\ \cline{3-6} 
                       &                         & Tokens/s       & Up             & Tokens/s       & Up             \\ \hline
0\%                    & Dense                   & 31.88          &                & 27.45          &                \\ \hline
\multirow{3}{*}{20\%}  & WandA                   & 32.05          & ×1.01          & 28.01          & ×1.02          \\
                       & FLAP                    & \textbf{38.90} & \textbf{×1.22} & \textbf{33.96} & \textbf{×1.24} \\
                       & SEAP-gen                & 37.32          & ×1.17          & 33.02          & ×1.20          \\ \hline
\multirow{3}{*}{50\%}  & WandA                   & 31.24          & ×0.98          & 27.01          & ×0.98          \\
                       & FLAP                    & \textbf{47.94} & \textbf{×1.50} & \textbf{43.45} & \textbf{×1.58} \\
                       & SEAP-gen                & 47.10          & ×1.48          & 41.78          & ×1.52          \\ \hline
\end{tabular}
\caption{Inference speed (Tokens per second) and speedup under different pruning ratios.A higher↑ speed indicates better performance.}
\label{tab:speed}
\end{table}
\vspace{-10pt}
\paragraph{Sparsity Setting Comparison}
As shown in Table~\ref{tab:sparsity}, we compare our Logistic-based (LB) sparsity setting with other strategies: Uniform Sparsity across Layers (UL) and Adaptive Sparsity across Layers and Modules (AL) from FLAP.
At both 20\% and 50\% pruning ratios, our method (SEAP-gen with LB) consistently outperforms WandA-sp and FLAP in terms of performance, demonstrating that the LB setting leads to more efficient resource allocation and better performance. 
\begin{table}[h]
\small
\setlength{\tabcolsep}{6.5pt}
\setlength{\belowcaptionskip}{0pt}
\begin{tabular}{@{}llcccc@{}}
\hline
Ratio                 & Method   & \multicolumn{1}{l}{Set.} & \multicolumn{1}{l}{Average} & \multicolumn{1}{l}{Set.} & \multicolumn{1}{l}{Average} \\ \hline
0\%                   & Dense    & -                        & 69.46                       & -                        & 69.46                       \\ \hline
\multirow{3}{*}{20\%} & WandA-sp & UL                       & 61.47                       & LB                       & \textbf{65.57}              \\
                      & FLAP     & AL                       & 63.03                       & LB                       & \textbf{66.76}              \\
                      & SEAP-gen & UL                       & 66.03                       & LB                       & \textbf{68.75}              \\ \hline
\multirow{3}{*}{50\%} & WandA-sp & UL                       & 48.80                       & LB                       & \textbf{49.94}              \\
                      & FLAP     & AL                       & 51.12                       & LB                       & \textbf{51.78}              \\
                      & SEAP-gen & UL                       & 59.03                       & LB                       & \textbf{60.89}              \\ \hline
\end{tabular}
\caption{Sparsity settings and average sparsity on the Llama-2-13B model. The table shows three sparsity strategies: "UL" (Uniform Layer Sparsity), "LB" (Logistic-based Sparsity), and "AL" (Adaptive Layer Sparsity).A higher↑ score indicates better performance.}
\label{tab:sparsity}
\end{table}
\vspace{-10pt}

\section{Related Works}
\label{sec: related_works}
The computational cost and inference time of LLMs significantly impact deployment. Researchers have addressed these challenges through model compression\cite{a16hbta_Michel_arxiv19,Zeroquant_Yao_NurIPS22,AWQ_Lin_GetMobile24}, quantization\cite{BinaryBERT_Bai_ACL21,OPTQ_Frantar_ICLR23}, structural modifications\cite{Mamba_Gu_arxiv23,RWKV_Peng_EMNLP23}, and optimized decoding. Sparsification has become a key technique, including Mixture of Experts (MoE) \cite{MoE_Shazeer_ICLR17}, which activates subsets of the network to improve efficiency while maintaining performance \cite{Baselayer_Lewis_ICML21,Gshard_Lepikhin_arxiv20,MoEfication_Zhang_ACL22}.

Pruning is another effective sparsification technique for reducing computational and memory costs, categorized into unstructured, structured, and activation pruning.
\textbf{Unstructured Pruning}, which sparsifies individual weights but can hinder hardware efficiency. Examples include SparseGPT \cite{SparseGPT_Frantar_arxiv23} and WandA \cite{wanda_Sun_ICLR24}.
\textbf{Structured Pruning}, which prunes entire units like channels or attention heads for improved hardware efficiency and inference speed, with methods like Bonsai \cite{Bonsai_dery_arxiv24}, QPruner \cite{QPruner_zhou_arxiv24}, LLM-Pruner \cite{LlmPruner_Ma_NeurIPS23}, FLAP \cite{flap_An_AAAI24}, and Depth2 \cite{depth2_li_arxiv24}.
\textbf{Activation Pruning} sparsifies network activations, reducing memory bandwidth during inference. Activation functions like SiLU and GeLU \cite{Relu1_Mirzadeh_arxiv23}, and variants like dReLU \cite{drelu_Song_arxiv24}, ReGLU \cite{reGLU_Raffel_JMLR20}, and RELU$^{2}$ \cite{primerrelu2_So_NeurIPS21, Relu2_Zhang_arxiv24} help reduce computational load. Methods like TEAL \cite{TEAL_liu_arxiv24}, CATS \cite{CATS_lee_arxiv24}, SCAP \cite{SCAP_chua_arxiv24}, QSparse \cite{QSparse_Wang_arxiv24}, and ProSparse \cite{ProSparse_Song_arxiv24} achieve training-free activation pruning.

\section{Conclusion}
\label{sec: conclusion}
We present \textbf{SEAP (Sparse Expert Activation Pruning)}, a training-free, task-adaptive pruning framework for LLMs, inspired by the clustering of hidden states and task-specific activation patterns. SEAP dynamically selects and activates the most relevant neurons for each task, reducing computational overhead while maintaining strong task performance. Extensive experiments demonstrate that SEAP significantly improves efficiency—outperforming baselines by over 20\% at 50\% pruning, while maintaining over 97.8\% of the original performance at 20\% pruning. These results highlight SEAP’s ability to achieve substantial sparsification with minimal performance degradation. By leveraging insights from hidden state clustering and activation-driven pruning, SEAP optimizes LLMs for real-world deployment, enabling more efficient, scalable, and adaptive language models. This approach paves the way for future advancements in structured pruning and task-aware model compression, making LLMs more accessible and practical across diverse applications.

\section*{Acknowledgments}

This work is supported by the National Natural Science Foundation of China under grant 62072463, the Scientific Research Fund of Renmin University of China (Central Universities Basic Scientific Research Funds) under project 24XNKJ61, and the Open Fund of the National Key Laboratory of Digital Publishing Technology, Founder Group. The corresponding author of this paper is Zhiyu Li.

\section*{Ethical Considerations}

This work introduces a pruning method for LLMs to improve efficiency, but it raises ethical concerns. Pruning decisions could unintentionally affect model performance or fairness on certain tasks. While our method aims to preserve task-specific performance, it is important to monitor its impact on fairness and utility, especially in critical applications. Furthermore, pruned models could have unintended consequences in domains requiring nuanced decision-making. Transparent deployment and ongoing evaluation are essential to address these concerns.

\section*{Limitations}

While SEAP improves inference efficiency, it does have some limitations. (1) Compared to other methods, our approach may result in a slight increase in perplexity, as it preserves task-specific parameters at the cost of some efficiency. (2) The acquisition of task-specific activation values could also benefit from more diverse data, and incorporating data synthesis techniques could improve model generalization. (3) Lastly, pairing SEAP with a simple task classifier to route tasks to the pruned model could further enhance efficiency, making the approach more adaptable in practical applications.

\bibliography{custom}

\clearpage

\appendix
\section{Experimental Settings}
\label{sec: appendix_exp}
\subsection{Task-Specific Corpus Construction}
In this study, we constructed a standardized task-specific corpus by reformatting the questions and answers from evaluation tasks into knowledge-rich inputs.

For each task, we began by extracting relevant components from the raw training data, including the question, answer options, and correct answers. These components were then formatted into standardized input prompts.
By combining the question, options, and correct answer into a unified input format, we provided the model with the full context of each task, as shown in Table\ref{tab:task_examples}. This structured input allows the model to learn task-specific patterns and understand the relationship between the question and the correct answer, ultimately improving its ability to make accurate predictions.

\renewcommand{\arraystretch}{1.25}
\begin{table*}[ht]
\small 
\setlength{\belowcaptionskip}{0pt}
\begin{tabular}{clcccccccc}
\hline
\multirow{2}{*}{\begin{tabular}[c]{@{}c@{}}Pruning\\ Ratio\end{tabular}} & \multicolumn{1}{c}{\multirow{2}{*}{Method}} & \multicolumn{8}{c}{Llama-2-13B}                                                                                                       \\ \cline{3-10} 
                                                                         & \multicolumn{1}{c}{}                        & WinoGrande     & OBQA           & HellaSwag      & PIQA           & ARC-c          & ARC-e          & BoolQ          & Average        \\ \hline
0\%                                                                      & Dense                                       & 72.14          & 45.20          & 79.37          & 80.52          & 48.98          & 79.42          & 80.58          & 69.46          \\ \hline
\multirow{6}{*}{20\%}                                                    & WandA-sp (\(s_W\))                             & 67.40          & 42.80          & 74.52          & 78.40          & {\ul 48.64}    & 76.73          & 70.49          & 65.57          \\
                                                                         & SEAP (\(s_W\))                                 & \textbf{71.98} & 43.60          & 78.73          & \textbf{80.69} & 48.46          & 77.61          & 74.68          & 67.96          \\
                                                                         & SEAP-gen (\(s_W\))                             & 69.85          & 43.20          & 78.13          & {\ul 80.47}    & 48.55          & \textbf{78.58} & 72.54          & 67.33          \\ \cline{2-10} 
                                                                         & FLAP (\(s_F\))                                 & 69.14          & 44.00          & 75.05          & 76.71          & 48.04          & 77.19          & {\ul 77.22}    & 66.76          \\
                                                                         & SEAP (\(s_F\))                                 & {\ul 70.64}    & \textbf{44.80} & \textbf{79.12} & \textbf{80.69} & 47.95          & 76.85          & 76.82          & {\ul 68.12}    \\
                                                                         & SEAP-gen (\(s_F\))                             & 70.09          & {\ul 44.20}    & {\ul 78.97}    & 80.09          & \textbf{50.17} & {\ul 78.37}    & \textbf{79.36} & \textbf{68.75} \\ \hline
\multirow{6}{*}{50\%}                                                    & WandA-sp (\(s_W\))                             & 53.51          & 37.20          & 46.77          & 66.97          & 35.24          & 60.14          & 49.76          & 49.94          \\
                                                                         & SEAP (\(s_W\))                                 & 58.96          & 40.60          & 66.91          & 76.77          & {\ul 44.03}    & {\ul 71.09}    & {\ul 57.43}    & 59.40          \\
                                                                         & SEAP-gen (\(s_W\))                             & {\ul 63.38}    & \textbf{44.40} & 66.75          & 76.55          & 43.43          & {\ul 71.09}    & 49.79          & 59.34          \\ \cline{2-10} 
                                                                         & FLAP (\(s_F\))                                 & 55.17          & 38.20          & 53.82          & 67.41          & 33.11          & 58.42          & 56.36          & 51.78          \\
                                                                         & SEAP (\(s_F\))                                 & \textbf{64.56} & 42.00          & \textbf{68.75} & {\ul 76.93}    & \textbf{45.05} & \textbf{71.84} & 52.57          & {\ul 60.24}    \\
                                                                         & SEAP-gen (\(s_F\))                             & 62.59          & {\ul 43.20}    & {\ul 67.05}    & \textbf{77.15} & 41.55          & 67.93          & \textbf{66.79} & \textbf{60.89} \\ \hline
\end{tabular}
\caption{Task performance accuracy on Llama-2-13B under different pruning ratios. A higher ↑ score indicates better performance. The \textbf{bolded} entries represent the highest scoring methods, while the \underline{underlined} entries represent the second highest scoring methods.}
\label{tab:llama2_13b}
\end{table*}

\subsection{Tasks}
For evaluating downstream task performance, we use the lm-eval harness\cite{eval-harness} to assess zero-shot performance across seven benchmark tasks. We ensured that all tools and datasets used are properly cited, comply with their licenses and intended uses, and meet ethical standards, including data privacy and documentation. These tasks test a wide range of natural language understanding challenges and include:

\begin{itemize}
    \item \textbf{BoolQ} \cite{boolq_clark_NAACL19}: Evaluates models' ability to answer yes/no questions based on context, testing comprehension and reasoning.
    \item \textbf{ARC Easy and ARC Challenge} \cite{arc_Peter_arxiv18}: Benchmarks from the AI2 Reasoning Challenge assessing reasoning on multiple-choice science questions; Easy set for direct retrieval, Challenge set for complex reasoning.
    \item \textbf{HellaSwag} \cite{hellaswag_zellers_ACL19}: Tests commonsense reasoning by having models select the most plausible continuation of a given sentence.
    \item \textbf{OBQA} \cite{obqa_Mihaylov_EMNLP18}: An open-book question answering task assessing models' ability to answer factual questions using a collection of documents.
    \item \textbf{PiQA} \cite{piqa_Bisk_AAAI20}: Focuses on physical commonsense reasoning, requiring models to select the correct solution from two choices for a given problem.
    \item \textbf{Winogrande} \cite{WinoGrande_Sakaguchi_arxiv19}: A large-scale dataset designed to evaluate models' ability to resolve commonsense reasoning tasks in the style of the Winograd Schema Challenge.
\end{itemize}

These tasks cover a broad spectrum of natural language understanding, from reasoning and commonsense knowledge to factual and situational understanding.

\begin{table*}[ht]
\centering
\begin{tabular}{l|p{12.5cm}}
\hline
\multicolumn{1}{l}{\textbf{Task}} & \textbf{Example Prompt} \\ \hline
\multirow{2.5}{*}{HellaSwag} & Then, the man writes over the snow covering the window of a car, and a woman wearing winter clothes smiles. Then, the man continues removing the snow on his car. \\ \cline{1-2} 
\multirow{2.5}{*}{PIQA} & How do I ready a guinea pig cage for its new occupants? Provide the guinea pig with a cage full of a few inches of bedding made of ripped paper strips, you will also need to supply it with a water bottle and a food dish. \\ \cline{1-2}
\multirow{2}{*}{OBQA} & The sun is the source of energy for physical cycles on Earth: plants sprouting, blooming, and wilting. \\ \cline{1-2}
\multirow{2}{*}{WinoGrande} & Katrina had the financial means to afford a new car while Monica did not, since Katrina had a high paying job. \\ \cline{1-2}
\multirow{2.5}{*}{ARC} & One year, the oak trees in a park began producing more acorns than usual. The next year, the population of chipmunks in the park also increased. Which best explains why there were more chipmunks the next year? Food sources increased. \\ \cline{1-2}
\multirow{3.5}{*}{GSM8K} & Natalia sold clips to 48 of her friends in April, and then she sold half as many clips in May. How many clips did Natalia sell altogether in April and May? Natalia sold 48/2 = 24 clips in May. Natalia sold 48 + 24 = 72 clips altogether in April and May. \\ \cline{1-2}
\multirow{2.5}{*}{BoolQ} & All biomass goes through at least some of these steps: it needs to be grown, collected, dried, fermented, distilled, and burned... Does ethanol take more energy to make than it produces? False \\ \cline{1-2}
\end{tabular}
\caption{Example Prompts from Various Tasks in the Task-Specific Corpus}
\label{tab:task_examples}
\end{table*}

\subsection{Baselines}
In this study, we select two representative methods as baseline models for comparison: Wanda and FLAP. Below is a detailed introduction to these two methods.
\paragraph{Wanda}\cite{wanda_Sun_ICLR24}
Wanda evaluates parameter importance by calculating the product of the weight magnitude and the \(\ell_2\)-norm of the corresponding input activation. It adopts a local pruning strategy, pruning weights associated with each output feature within a linear layer. We extend Wanda to structured pruning by computing the \(\ell_2\)-norm of weight groups within the linear layer, evaluating the importance of the entire group. This extended version, called Wanda-sp, enables structured pruning in large language models.

\paragraph{FLAP}\cite{flap_An_AAAI24}
FLAP (Fluctuation-based Adaptive Structured Pruning) is a novel structured pruning method for large language models, achieving compression without retraining. It uses a fluctuation pruning metric to assess the recoverability of the output feature map after removing a column of weights. By normalizing importance scores, FLAP adaptively determines the global structure of the compressed model.

\subsection{Hyperparameters}
The hyperparameters in this study involve the weighting of tasks and the sparsity setting across layers.

For general pruning, the importance score \( s_i^{(\ell)} \) for each neuron in layer \(\ell\) is calculated as a weighted sum of task-specific scores:
\[
s_i^{(\ell)} = \sum_{\tau} \alpha_{\tau} s_i^{(\ell,\tau)},
\]
where \(\alpha_{\tau}\) is the weight assigned to task \(\tau\). WikiText2 is assigned a weight of 3 as an expert activation for language modeling, while other tasks are assigned an equal weight of 2.

To achieve a global sparsity target \(G\), we adjust the sparsity distribution across layers by tuning the parameter \(\Lambda\) such that the average sparsity satisfies:
\[
\frac{1}{L} \sum_{\ell=1}^{L}\rho_{\ell} = G.
\]
This is done through a numerical search for the optimal \(\Lambda\). In our experiments, we use \((x_0, k) = (0.3, 1)\) for the logistic sparsity function.

\begin{figure}[t]
    \centering
    \includegraphics[width=\linewidth]{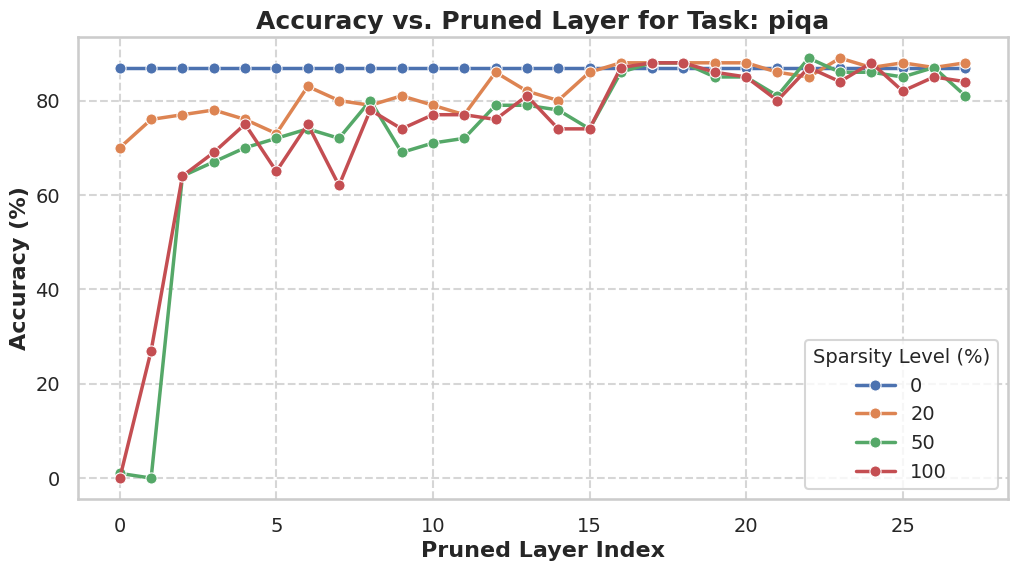}
    \caption{Impact of pruning on PIQA performance at different layers and sparsity levels. Deeper layers are more robust to pruning.}
    \label{fig:piqa_rm_test}
\end{figure}

\section{Additional Experiments}
\label{subsec: appendix_add}

In this section, we present two additional experiments to support our proposed method. These experiments are designed to assess key aspects of the model’s performance: perplexity as a measure of language modeling quality and task classification for task-specific pruning.

\subsection{Perplexity}
\label{subsec: perplexity}

We evaluate the impact of pruning on language modeling by assessing perplexity (PPL) on the WikiText2 dataset. Perplexity measures how well a model predicts the next word in a sequence, with lower values indicating better performance. This experiment helps determine whether pruning methods, including SEAP, can maintain language generation quality while achieving computational savings.

\begin{figure}[t]
    \centering
    \includegraphics[width=\linewidth]{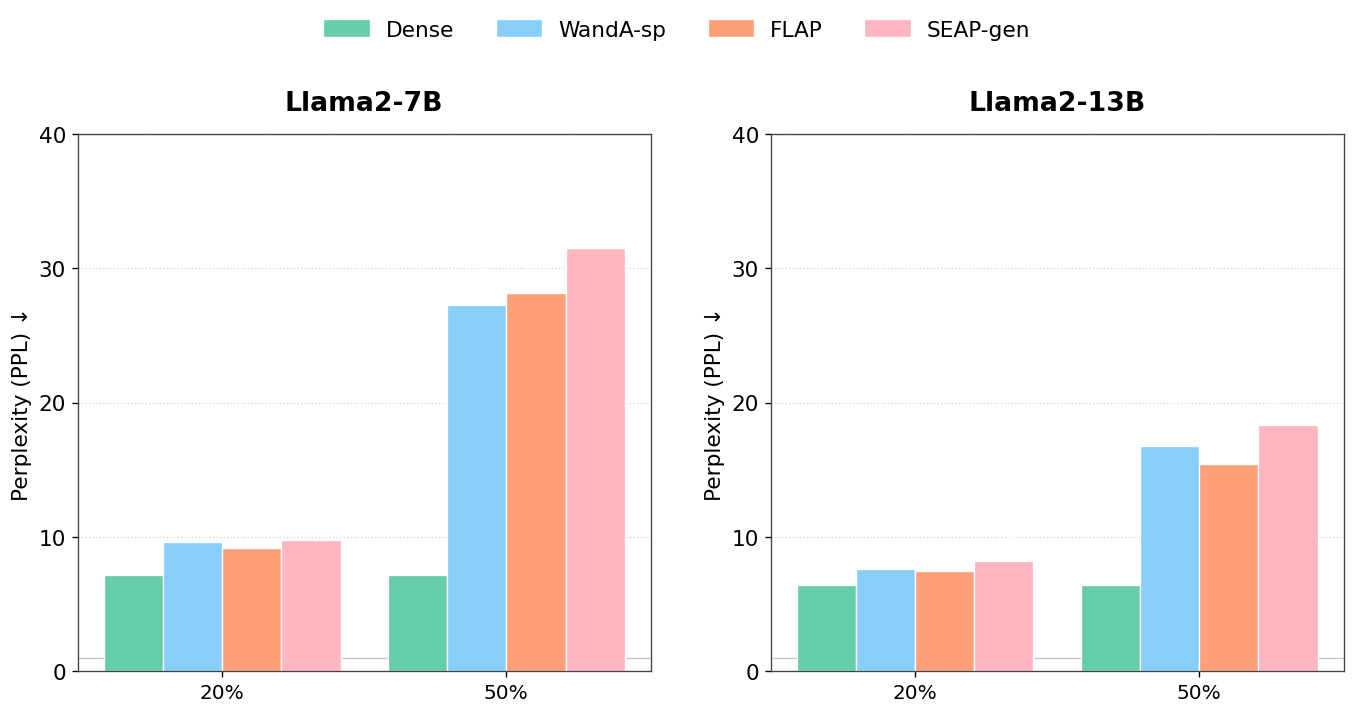}
    \caption{Perplexity (PPL) results under different pruning ratios. A lower↓ perplexity indicates better performance.}
    \label{fig:ppl}
\end{figure}

We use 128 random samples from the WikiText2 dataset \cite{wikitext2_Merity_arxiv16}, each with a 2048-token context and a 512-token evaluation window, following the FLAP setup \cite{flap_An_AAAI24}. As shown in Figure~\ref{fig:ppl}, at 20\% sparsity, SEAP leads to a slight increase in perplexity compared to WandA-sp and FLAP, reflecting a small trade-off in language modeling quality. At 50\% sparsity, perplexity increases across all methods, with SEAP-gen showing the highest values. However, these increases remain within an acceptable range, especially considering the significant improvements in task-specific performance.

\subsection{Task Classifier}
\label{subsec: task_classifier}

A key feature of the task-specific expert activation pruning method is its ability to dynamically select pruning masks based on the task type, improving computational efficiency. The challenge lies in quickly identifying the task type with minimal overhead to ensure efficient mask selection.

To address this, we propose a lightweight task classification method. We extract a vector from the model's 0th-layer embedding and train a single-layer classifier to identify the task type, enabling the model to select the appropriate task-specific mask with minimal cost.

\begin{table}[h!]
\centering
\small
\begin{tabular}{lp{0.92cm}p{0.92cm}p{0.92cm}p{0.92cm}}
\hline
\textbf{Class}        & \textbf{Precision} & \textbf{Recall} & \textbf{F1-Score} & \textbf{Support} \\ \hline
hellaswag             & 0.94               & 0.89            & 0.91              & 236              \\
gsm8k                 & 0.92               & 0.99            & 0.95              & 233              \\
winogrande            & 0.98               & 0.99            & 0.98              & 218              \\
piqa                  & 0.88               & 0.91            & 0.89              & 201              \\
mmlu                  & 0.95               & 0.87            & 0.91              & 233              \\
ai2\_arc              & 0.91               & 0.95            & 0.93              & 222              \\ \hline
\textbf{Accuracy}     & \multicolumn{4}{c}{0.93 (1343)}                                             \\ \hline
\textbf{Macro avg}    & 0.93               & 0.93            & 0.93              & 1343             \\ \hline
\textbf{Weighted avg} & 0.93               & 0.93            & 0.93              & 1343             \\ \hline
\end{tabular}
\caption{Task classification performance metrics for the proposed task classifier. Precision, recall, and F1-score are reported for each task class, along with the overall accuracy, macro average, and weighted average.}
\label{tab:task_classification}
\end{table}

As shown in Table~\ref{tab:task_classification}, the task classifier performs effectively, distinguishing between different task types with minimal computational resources. This confirms that task classification can be done rapidly, allowing for efficient pruning and dynamic adjustment of model parameters.

\section{Generation Examples}
\label{sec: appendix_example}

In this section, we provide several examples of generated outputs to illustrate the effectiveness of our method. The following tables showcase two results of our model's generation capabilities using prompts from the FLAP: Table~\ref{tab:llama2_7b_examples} and Table~\ref{tab:llama2_13b_examples}.

\begin{table*}[ht]
\centering
\begin{tabular}{lp{0.8\textwidth}} 
\hline
\multicolumn{1}{l}{\textbf{Model}}                                              & \textbf{Generated Sentences}                                                                                                                                                                                                                                                                                                                                                                                             \\ \hline
\multirow{8}{*}{\begin{tabular}[c]{@{}c@{}}LLaMA-7B\\ 20\% Pruned\end{tabular}} & AI can create a logo in seconds. But it’s not just a logo that you need to build a brand. A brand is a collection of ideas, emotions, and associations that people have about a product, service, or company. It’s the sum of all the things that make up a brand, and it’s what differentiates one brand from another. The most important thing to remember about branding is that it’s not just about the logo.        \\ \cline{2-2} 
                                                                                & What is McDonald's? McDonald's is a fast-food restaurant franchise that serves burgers, fries, chicken nuggets, soft drinks, milkshakes, sundaes, salads, wraps, breakfast items (in some restaurants), desserts, and other items. The menu varies by country and region, but there are over 30,000 locations worldwide.                                                                                                 \\ \hline
\multirow{8}{*}{\begin{tabular}[c]{@{}c@{}}LLaMA-7B\\ 50\% Pruned\end{tabular}} & AI can create a logo in seconds. The process of creating logotypes is called typestudy. A Log is a handwritten symbol or group of alphabets used in a script. In 1915, Eric Blake created the typeface Helvetica. It was a revival of handwriting from the Italian Renaissance. The first known example of handwriting is from a cave in Georgia, USA. It is dated 10,000 BCE.                                           \\ \cline{2-2} 
                                                                                & What is McDonald's? It is a small inn in an alley, a hundred yards or two from the gates. The tavern's walls are thick, and it has a steeply pitched roof. Above the door there is carved a dragon's flicked beak, with the words \_Bow of Arrows\_ carved in the sill. Inside, the tavern is large and well lit by daylight. There are three stairways leading to the north, all of which are occupied by men in armor. \\ \hline
\end{tabular}
\caption{Generated Sentences by LLaMA-7B with Different Pruning Levels}
\label{tab:llama2_7b_examples}
\end{table*}

\begin{table*}[ht]
\centering
\begin{tabular}{lp{0.8\textwidth}} 
\hline
\multicolumn{1}{l}{\textbf{Model}}                                              & \textbf{Generated Sentences}                                                                                                                                                                                                                                                                                                                                                                                             \\ \hline
\multirow{8}{*}{\begin{tabular}[c]{@{}c@{}}LLaMA-13B\\ 20\% Pruned\end{tabular}} & "AI can create a logo in seconds. But it’s not that simple. There are pros and cons to AI-generated logos. On the one hand, they are quick to create and don’t require a lot of time or resources. On the other hand, they can be generic and lack personality. In this article, we’ll explore the pros and cons of AI-generated logos and how they compare to human-designed logos. " \\ \cline{2-2} 
                                                                                & "What is McDonald's? McDonald's is a fast-food hamburger restaurant that serves burgers, fries, chicken nuggets, milkshakes, salads, and breakfast items. The menu is simple and affordable, and the restaurant is known for its cleanliness and friendliness. What is the history of McDonald's? In 1940, two brothers, Ray Kroc and Richard McDonald, opened the first McDonald's restaurant in Des Plaines, Illinois." \\ \hline
\multirow{11}{*}{\begin{tabular}[c]{@{}c@{}}LLaMA-13B\\ 50\% Pruned\end{tabular}} & "AI can create a logo in seconds. It’s a matter of fact that the time to create a logo has decreased from 10 days to 24 hours in the last decade. This is due to the development of computer graphics and digital technologies. In the 1990s, the world’s first computer-animated film “Turtle Island” was released in 1990. It took three years to make the film and cost \$40 million. The sequel of the franchise, Taz the Stone Age, was released in 1994. It made \$ 24 million in box office and gross revenue of \$402 million worldwide." \\ \cline{2-2} 
                                                                                & "What is McDonald's? Founded in 1946 by Mac and his brother Dave McDonald in Aberdeen, Scotland as a bar for American servicemen and their friends, the Macdonald family took over the business in 1972. It was renamed The White Rose in 1974 and changed to its current name in 1986 due to the unfortunate similarity of McDonald's which was registered at the time some 30 miles away. Today the pub serves 1200 pints a week and has live music 6 nights a week with DJ's playing between breaks." \\ \hline
\end{tabular}
\caption{Generated Sentences by LLaMA-13B with Different Pruning Levels}
\label{tab:llama2_13b_examples}
\end{table*}

\end{document}